# Face Identification by SIFT-based Complete Graph Topology

Dakshina Ranjan Kisku, Ajita Rattani
Department of Computer Science and Engineering
Indian Institute of Technology Kanpur
Kanpur, India
{drkisku, ajita}@iitk.ac.in

Enrico Grosso, Massimo Tistarelli
Computer Vision Laboratory
University of Sassari
07100 Sassari , Italy
{grosso, tista}@uniss.it

***Abstract*—This paper presents a new face identification system based on Graph Matching Technique on SIFT features extracted from face images. Although SIFT features have been successfully used for general object detection and recognition, only recently they were applied to face recognition. This paper further investigates the performance of identification techniques based on Graph matching topology drawn on SIFT features which are invariant to rotation, scaling and translation. Face projections on images, represented by a graph, can be matched onto new images by maximizing a similarity function taking into account spatial distortions and the similarities of the local features. Two graph based matching techniques have been investigated to deal with false pair assignment and reducing the number of features to find the optimal feature set between database and query face SIFT features. The experimental results, performed on the BANCA database, demonstrate the effectiveness of the proposed system for automatic face identification.**

## I. INTRODUCTION

During the past decade, face identification and verification has drawn significant attention from the perspective of different real life applications such as human computer interface, surveillance, authentication and video indexing. Due to variations in illumination, nearby clutter, variability in scale, translation, rotation, and pose, face identification is a challenging task. Facial expression, occlusion and lighting conditions also change the overall appearance of the face.

Many efforts have been devoted to solve the threats owing to face identification systems, which result in the severe degradation of the performance. Although many appearance-based face identification or verification techniques based on the component analysis such as in [1], [9] and [10] exist in the literature, they are inefficient to capture a substantial amount of facial variations or new class samples. Reference [11] proposed a face recognition system by elastic bunch graph matching technique. However, the performance of the system has not been tested under different constraints and furthermore, the overall identification process resulted quite complicated. Reference [3] has proposed a probabilistic face recognition approach that could compensate for the imprecise localization, partial occlusion, and extreme expressions with a single training sample. However, the illumination problem remained unsolved. Of late, the investigation of SIFT features for face authentication has been explored in [8]. The results were very promising; but the need is felt for more robust matching techniques to further improve the overall system performance.

In the proposed method the face image is first photometrically normalized by using histogram equalization. The rotation, scale and translation invariant SIFT features are extracted from the face image. Finally the graph-based topology is used for matching two face images. Two matching techniques are applied: the gallery image based and the reduced point based match constraint. The results are obtained from the BANCA database using the MC protocol.

The remainder of the paper is organized as follows: section 2 briefly describes the affine invariant and robust SIFT features extraction process. Section 3 describes the taxonomy of correspondence graph used as a matcher and the proposed SIFT generated graph matching techniques. Experimental results are given in section 4 and in the last section the conclusions are drawn.

## II. INVARIANT AND ROBUST SIFT FEATURES

In object recognition and image retrieval applications, affine-invariant features have been recently researched [5], [6]. These affine-invariant features are highly distinctive and matched with high probability against a large case of image distortions and illumination conditions.

The scale invariant feature transform, called SIFT descriptor, has been proposed by [2], [4] and proved to be invariant to image rotation, scaling, translation, partly illumination changes, and projective transform. The basic idea of the SIFT descriptor is detecting feature points efficiently through a staged filtering approach that identifies stable points in the scale-space.

Local feature points are extracted from the following steps:

- select candidates for feature points by searching peaks in the scale-space from a difference of Gaussian (DoG) function,
- localize the feature points by using the measurement of their stability,

- assign orientations based on local image properties,
- calculate the feature descriptors which represent local shape distortions and illumination changes.

After candidate locations have been found, a detailed fitting is performed to the nearby data for the location, edge response, and peak magnitude. To achieve invariance to image rotation, a consistent orientation is assigned to each feature point based on local image properties and describes it relative to this orientation. The histogram of orientations is formed from the gradient orientation at all sample points within a circular window of a feature point. Peaks in this histogram correspond to the dominant directions of each feature point.

For illumination invariance 8 orientation planes are defined. Towards this end, the gradient magnitude and the orientation are smoothed by applying a Gaussian filter and then sampled over a 4 x 4 grid with 8 orientation planes.

## III. Representation of Faces

In this work each face is represented with a complete graph drawn on feature points extracted using the SIFT operator [4]. Two matching constraints are proposed: gallery image based match constraint and reduced point based match constraint. These techniques can be applied to find the corresponding sub-graph in the probe face image given the complete graph in the gallery image.

### *A. Taxonomy of correspondence graph in the context of graph matching constraint*

Let G1 and G2 are two face image graphs given by:

$$G1 = \{V^{G1}, E^{G1}, F^{G1}\},\ G2 = \{V^{G2}, E^{G2}, F^{G2}\} \qquad (1)$$

where $V^{Gk}$, $E^{Gk}$ and $F^{Gk}$ represent the list of vertexes, edges and SIFT features associated to the graph, where k is either 1 or 2 for representation of two face images.

Let us define the directional correspondence between two nodes as follows:

*1) Definition 1:* The i<sup></sup>th (with i = 1,2,……,N) feature point in the face graph G1 has correspondence to the j<sup></sup>th (with j = 1,2,…….,M) feature point in the face graph G2 in terms of conditional probability $V_i^{G1} \rightarrow V_j^{G2}$, if $| p(V_i^{G1} = V_j^{G2}, V_{i+1}^{G1} = V_j^{G2}, \ldots) \mid G1) — 1 | \leq \epsilon;\ \epsilon > 0$. Note that $V_i^{G1} \rightarrow V_j^{G2}$ does not imply $V_j^{G2} \rightarrow V_i^{G1}$. Therefore, one-to-one correspondence as the extension of the directional correspondence is defined to omit the false matches.

*2) Definition 2:* The $i^{th}$ feature point in the Face Graph G1 and the $j^{th}$ feature point in the face graph G2 have one-to-one correspondence in terms of conditional probability, if $| p(V_i^{G1} = V_j^{G2}) \mid G1) — 1 | \leq \epsilon_1$ and $| p(V_j^{G2} = V_i^{G1}) \mid G2) — 1 | \leq \epsilon_2$ for small $\epsilon_1 > 0$, $\epsilon_2 > 0$ and denoted by $V_i^{G1} \leftrightarrow V_j^{G2}$.

The correspondence graph between G1 and G2 by is defined as:

$$G^{G1\leftrightarrow G2} = (V^{G1}, V^{G2}, E^{G1}, E^{G2}, F^{G1}, F^{G2}, C^{G1\leftrightarrow G2}) \qquad (2)$$

where $\overline{G}k \subseteq Gk$, k = 1,2 are sub-graphs of the original graphs, in which all nodes have the one-to-one correspondence to each other, such that $V_i^{G1} \leftrightarrow V_j^{G2}$.

$C^{G1\leftrightarrow G2}$ is the set of node pairs, which have the one-to-one correspondence, given by

$$C^{G1\leftrightarrow G2} = \{(V_i^{G1}, V_j^{G2}) \mid V_i^{G1} \leftrightarrow V_j^{G2}\} \qquad (3)$$

Based on these two definitions, two match constraints are developed and thoroughly investigated.

### *B. Graph Matching Methodologies*

As explained in section 2, each feature point is composed by four types of information: spatial coordinate, key point descriptor, scale and orientation. Key point descriptor is a vector of 1x128 values. For sake of representation, given a feature point $f_i$, let's consider $X(f_i)$, $K(f_i)$, $S(f_i)$ and $O(f_i)$ its spatial coordinate or location, the key point descriptor, scale and orientation, respectively.

In the following sections the different graph matching methodologies are described.

*1) Gallery image based match constraint:* It is assumed that matching points will be found around similar positions i.e., fiducial points on the face image.

To eliminate false matches a minimum Euclidean distance measure is applied. It may be possible that more than one point in the first image correspond to the same point in the second image. Let's consider, $X_1, X_2, \ldots, X_N$ are the interest points found in the first image and $Y_1, Y_2, \ldots, Y_M$ are the interest points found in the second image. Here, $N \leq M$ or $N \geq M$ (N = number of interest points on the first image; M = number of interest points on the second image), both the cases would be possible.

Whenever $N \leq M$, many interest points from the second image are discarded, while if $N \geq M$, many repetitions of the same point would be occurred as corresponding points in the second image.

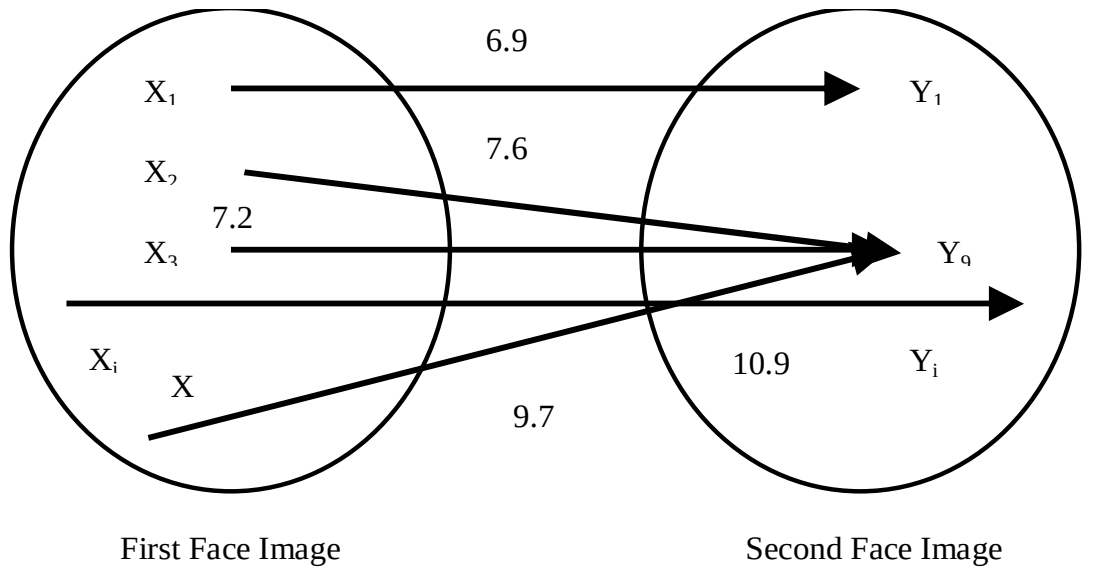

Figure 1. The corresponding points of image 1 are mapped into image 2 using the minimum Euclidean distance measure.

In both cases one interest point of the second image may correspond to several points in the first image. After computing all the distances, only the point with the minimum distance from the corresponding point in the second image is paired (see figure 1 and 2).

The distances are computed as the Hausdorff distance between two images usually computes. The dissimilarity scores are computed between all pairs of nodes of two face images after constructing the complete graph for each face image.

Given two images $F_{gallery}$ and $F_{probe}$, representing the gallery and probe image, respectively, and I is the compact information composed of all the four types of information generated by the SIFT descriptor, than:

$$I(F_{gallery}) = \{X^{F_{gallery}}(f_i), K^{F_{gallery}}(f_i), S^{F_{gallery}}(f_i), O^{F_{gallery}}(f_i)\}$$
$$I(F_{probe}) = \{X^{Fprobe}(f_j), K^{Fprobe}(f_j), S^{Fprobe}(f_j), O^{F_{probey}}(f_j)\}$$

where i = 1,2,…….,N; j = 1,2,………,M;

The dissimilarity scores $D^{GIBMC}{}_V(F_{gallery}, F_{probe})$ and $D^{GIBMC}{}_E(F_{gallery}, F_{probe})$ are computed for both the vertexes and the edges as:

$$D^{GIBMC}{}_V(F_{gallery}, F_{probe}) = \min_{i=1,2,\ldots,N}\{\min_{j=1,2,\ldots,M}(d(I_i(F_{gallery}), I_j(F_{probe}))\} \quad (4)$$

$$\Delta^{GIBMC}{}_V(F_{gallery}, F_{probe}) = \frac{1}{T}\sum_{t=1}^{T} D^{GIBMC}{}_V(F_{gallery}, F_{probe})^t \quad (5)$$

where T the total number of all is minimum distances. To find the correspondences between two graphs in terms of edge information, let’s take $E_1(N)$ and $E_2(N)$ are the number of edges in the two face graphs, respectively. Here, the number of nodes for image one is same as image two. After finding the corresponding points between the feature points of the first and the second image, we construct complete graph for each face image. Now, we try to find out the correspondences and compute the dissimilarity scores for a pair of edges. If nodes $a, a' \in N^{gallery}$ and $b, b' \in N^{probe}$ belongs to gallery and probe images, respectively, then $(a, a') \in E_1(N)$ and $(b, b') \in E_2(N)$ would be a pair of edges for the gallery image and the probe image, respectively.

$$D^{GIBMC}{}_E(F_{gallery}, F_{probe}) = d(I_i(F_{gallery}(a, a')), I_j(F_{probe}(b, b'))) \quad (6)$$

where i,j = 1,2,……….,N;

$$\Delta^{GIBMC}{}_E(F_{gallery}, F_{probe}) = \frac{1}{E}\sum_{e=1}^{E} D^{GIBMC}{}_E(F_{gallery}, F_{probe})^e \quad (7)$$

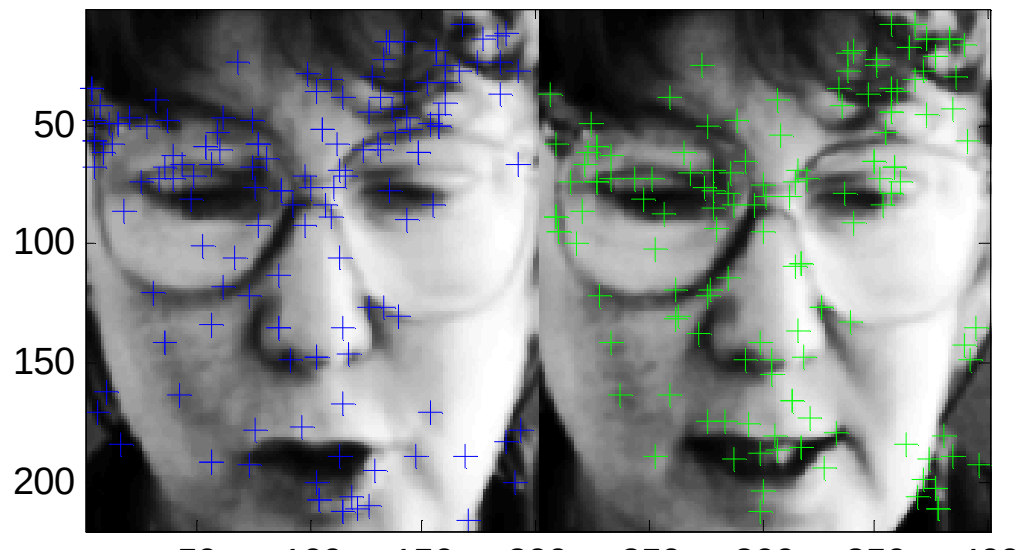

Figure 2. All the feature points and their matches for a pair of images, based on the Euclidean distance measure.

where $d(I_i(F_{gallery}(a, a')), I_j(F_{probe}(b, b')))$ is the distance between a pair of edges and $D^{GIBMC}{}_E(F_{gallery}, F_{probe})$ is the average distance of all pairs of edges.

*2) Reduced point based match constraint:* After eliminating the points that do not satisfy the gallery image point based match constraints, there can still be some false matches. Usually, the false matches are due to multiple assignments, which exist when more than one point (e.g, X2, X3 and XN) are assigned to a single point (e.g. Y9) in the other image. They can be also due to one way assignments, which exist when a point X2 is assigned to a point Y9 on the other image while the point Y9 is assigned to other two points X3 and XN or not assigned to any point (see figure 3).

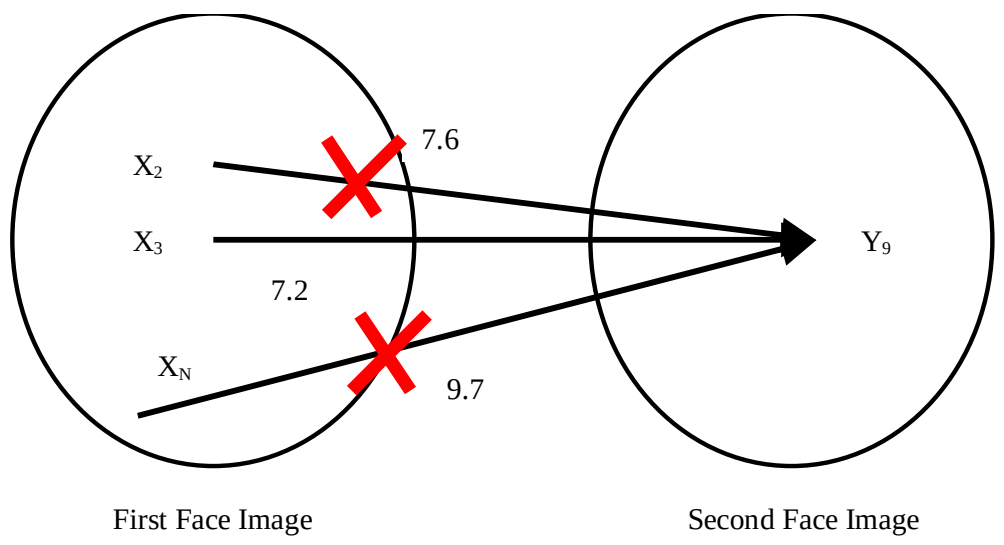

Figure 3. For a pair of faces X and Y, let X2, X3 and XN be three points on X; and Y9 is a point on Y with the arrows showing the matches and their direction. On the left is a multiple assignment where points X2, X3 and XN on X match Y9 on Y. In such case, the match between X2 and Y9, while XN and Y9 are eliminated.

These false matches can be eliminated with the application of another constraint, namely the reduced point based match constraint, which guarantees that each assignment from an image $X$ to another image $Y$ will have a corresponding assignment from image $Y$ to image $X$. With this constraint, the false matches due to multiple assignments are eliminated by choosing the match with the minimum distance. The false matches due to one way assignments are eliminated by removing the links which do not have any corresponding assignment from the other side. Examples showing the matches before and after applying the reduced point based match constraints are given in Figure 4(a) and 4(b).

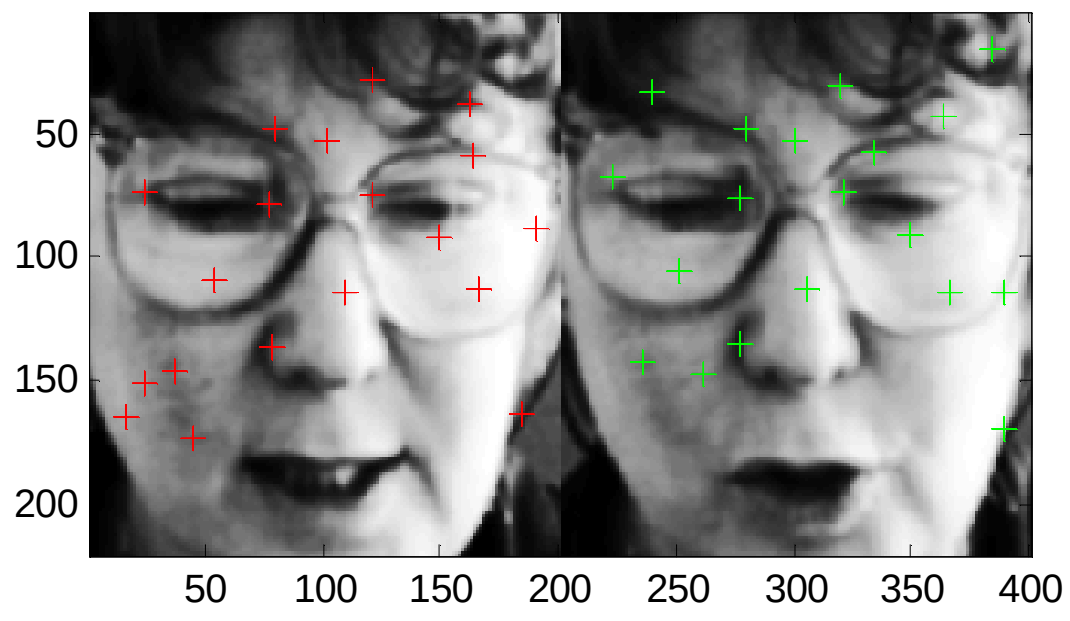

4(a)

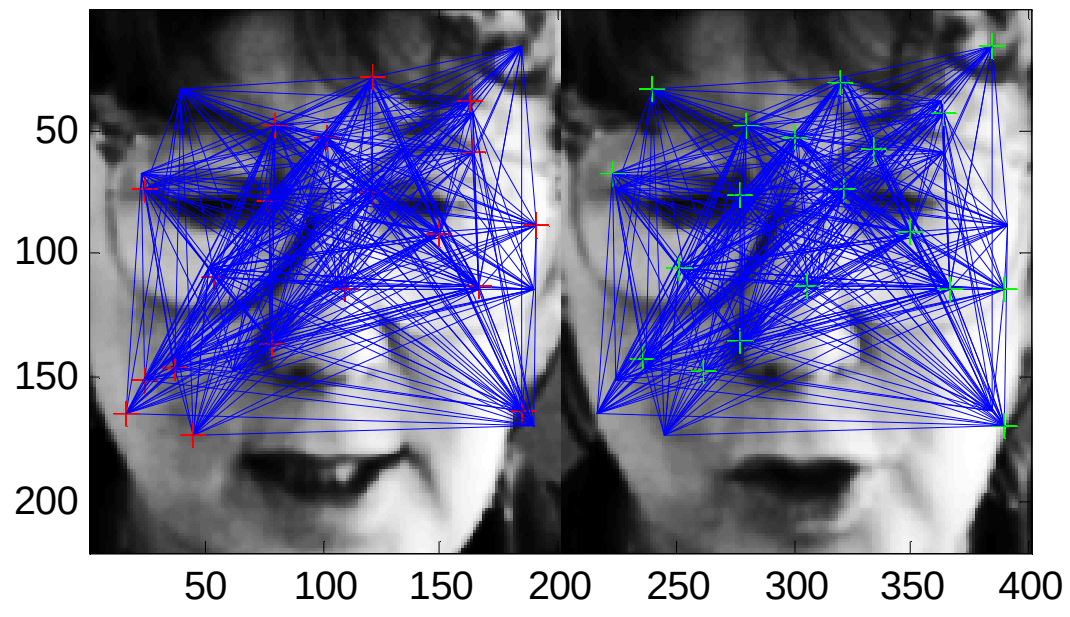

4(b)

Figure 4. An example of reduced point based match constraint. (a) All matches computed from the left to the right image. (b) The resulting complete graphs with a few numbers of false matches.

In this graph matching strategy, the same approach is followed which is described in section 3.B.1 for the gallery image based match constraint. False matches, due to multiple assignments, are removed by choosing the match with the minimum distance between two face images. The dissimilarity scores on reduced points between two face images for nodes and edges, is computed in the same way as for the gallery based constraint.

Lastly, the average weighted score is computed. This graph matching technique is more efficient than gallery image based match constraint, since the matching is done on a very small number of feature points with very few floating feature points.

*3) Weighting the score reliability:* The quality of the features has a significant impact on the performance of any learning based recognition algorithm. How to improve the quality of features has been one of the critical issues concerned with the instance-based learning.

Various approaches have been proposed in the past to address this issue. These approaches can be mainly divided into feature selection and feature weighting [12].

This work proposes a feature weighting method that is based on the Gaussian empirical rule. In this method, the relevance of a feature is determined by assigning a weight using the Gaussian empirical rule. The rationale behind this idea is that a relevant feature should have strong impact on classification. One advantage of using the Gaussian empirical rule for feature weighting is its rich expressiveness in representing hypotheses. In order to determine the weighted distance between two graphs, the weights value can be assigned by applying the Gaussian empirical rule in which three properties defined:

- 68% of the observations fall within 1 standard deviation of the mean, i.e. between $\mu$-$\sigma$ and $\mu$+$\sigma$.
- 95% of the observations fall within 2 standard deviations of the mean, i.e. between $\mu$-2$\sigma$ and $\mu$+2$\sigma$.
- 99.7% of the observations fall within 3 standard deviations of the mean, i.e. between $\mu$-3$\sigma$ and $\mu$+3$\sigma$.

Before generating the weighted dissimilarity scores, for each pair of face images, the mean and standard deviation are computed for a set of nodes and for a set of edges on the face graph.

If the node and edge values lie within one, two and three times the standard deviation of the mean, they are multiplied by 0.075 or 0.05 or 0.025 respectively. These values have been determined by a through testing on the BANCA database.

## IV. Experimental Results

The proposed graph matching technique is tested on the BANCA database [7]. For this experiment, the Matched Controlled (MC) protocol is followed, where the images from the first session are used for training, whereas second, third, and fourth sessions are used for testing and generating client and impostor scores. The testing images are divided into two groups, G1 and G2, of 26 subjects each. The error rate was computed using the following procedure [7]:

- For getting G1 scores, perform the experiment on G1.
- Perform the experiment on G2, getting G2 scores.
- Compute the ROC curve using G1 scores; determine the Prior Equal Error Rate and the corresponding client-specific threshold for each subject or each individual from several instances.
- Use the threshold $T_{G1}$ to compute False Acceptance Rate ($FAR_{G2}(T_{G1})$) and False Rejection Rate ($FRR_{G2}(T_{G1})$) on the G2 scores. The threshold is client-specific i.e, computed specifically for each individual from the several instances of his/her images.
- Compute the weighted Error Rate (WER(R)) on G2:

$$WER(R) = \frac{FRR_{G2}(T_{G1}) + R \times FAR_{G2}(T_{G1})}{1 + R} \quad (8)$$

for R = 0.1, 1 and 10;

WER(R) is computed on G1 by a dual approach, where the parameter R indicates the cost ratio between false acceptance and false rejection.

Prior Equal Rates for G1 and G2 are presented in table 1 and 2 showing the weighted equal error rates for three different values of R. The corresponding ROC curves are shown in Figure 5.

From the results presented the advantage of taking into account the context information in form of complete graph topology can be easily understood.

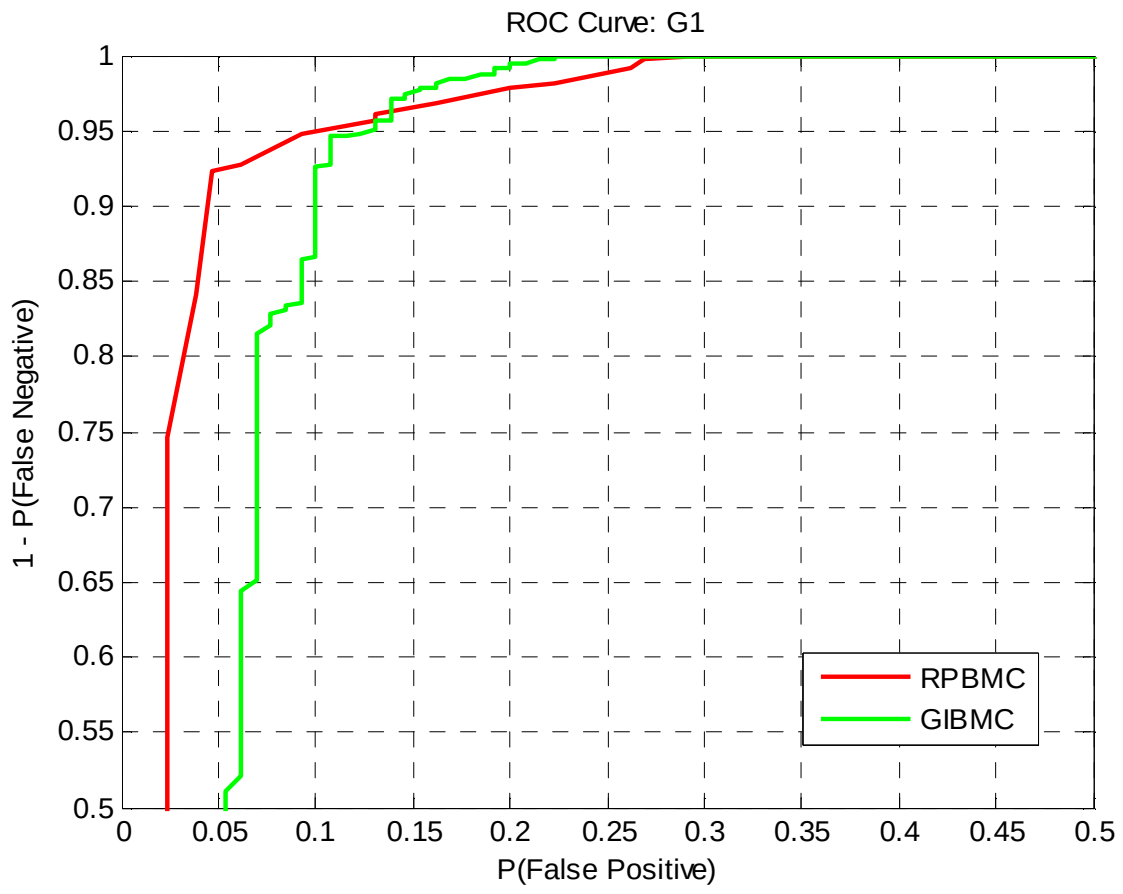

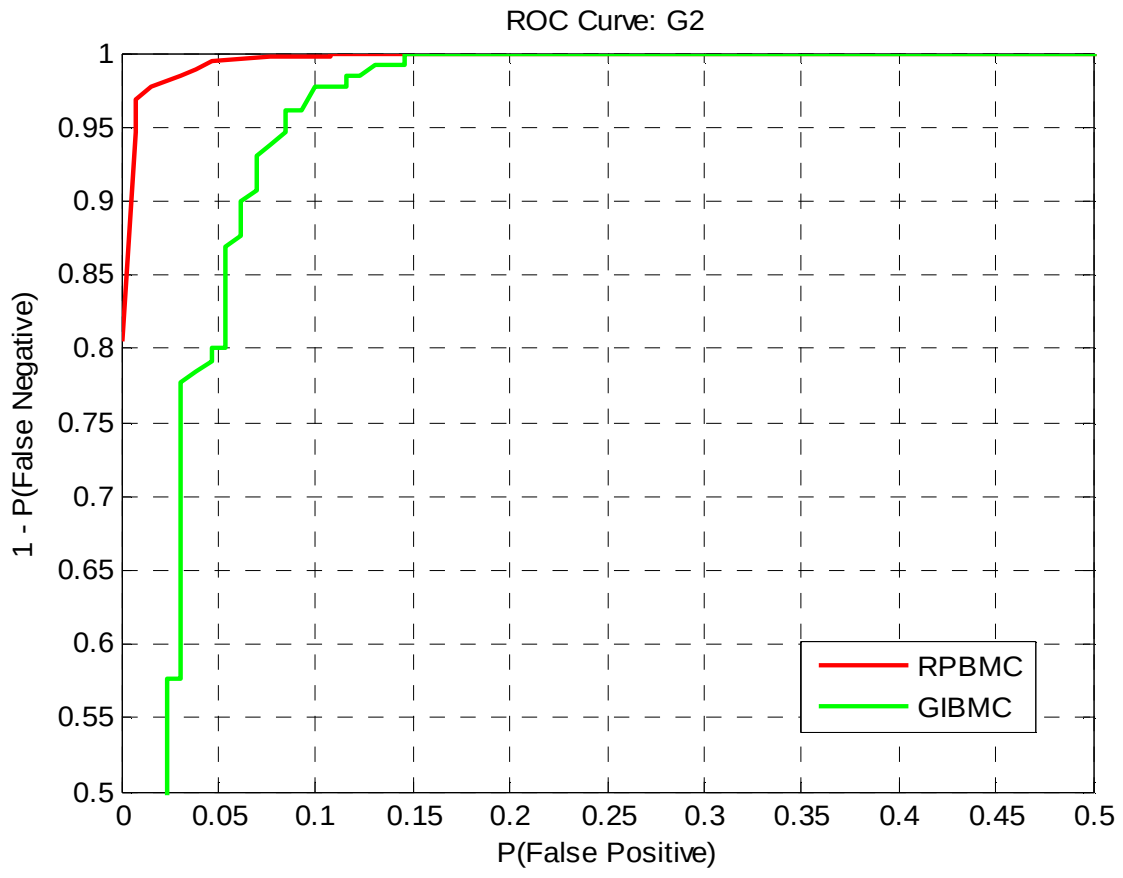

Figure 5. ROC curves for G1 (top) and G2 (bottom): 'GIBMC' stands for *Gallery Image Based Match Constraint*, 'RPBMC' stands for *Reduced Point Based Match Constraint*.

TABLE I. PRIOR EER ON G1 AND G2 FOR THE TWO METHODS: 'GIBMC' STANDS FOR *GALLERY IMAGE BASED MATCH CONSTRAINT*, 'RPBMC' STANDS FOR *REDUCED POINT BASED MATCH CONSTRAINT*

| | **GIBMC** | **RPBMC** |
|---|---|---|
| **Prior EER on G1** | 10.13% | 6.66% |
| **Prior EER on G2** | 6.92% | 1.92% |
| **Average** | 8.52% | 4.29% |

TABLE II. WER FOR THE TWO DIFFERENT GRAPH MATCHING TECHNIQUES: 'GIBMC' STANDS FOR *GALLERY IMAGE BASED MATCH CONSTRAINT*, 'RPBMC' STANDS FOR *REDUCED POINT BASED MATCH CONSTRAINT*

| | **GIBMC** | **RPBMC** |
|---|---|---|
| WER(R=0.1) on G1<br>WER(R=0.1) on G2 | 10.24%<br>6.83% | 7.09%<br>2.24% |
| WER(R=1) on G1<br>WER(R=1) on G2 | 10.13%<br>6.46% | 6.66%<br>1.92% |
| WER(R=10) on G1<br>WER(R=10) on G2 | 10.02%<br>6.09% | 6.24%<br>1.61% |

## V. CONCLUSION

This paper proposes two methods for face identification based on the SIFT [4] to generate a complete graph representation. The database and query face images are matched by finding the corresponding feature points using two matching constraints to deal with false pair assignments and optimal feature sets. The results are obtained by testing the methods on the BANCA database using the MC. This work shows a remarkable increase in the performance of the system with respect to the previous work based on the SIFT features. The obtained results show the capability of the system to cope for illumination changes and occlusions occurring in the database or the query face image. The technique presented can be compared with the elastic bunch graph matching technique [11] which is based on a straightforward comparison of image graphs. Identification experiments with the EBGM are reported on the FERET database as well as the Bochum database, including recognition across different poses, but the reported errors are higher than those obtained from the presented system. For the future, further tests will be performed to allow a direct comparison of the results from the two methods.

The reported identification accuracy further highlights that improved matching techniques can increase the performance of the system even in the same feature representation space.

## ACKNOWLEDGMENT

This work has been partially supported by grants from the Italian Ministry of Research, the Ministry of Foreign Affairs and the Biosecure European Network of Excellence. The authors would also like to thank Dr Manuele Bicego and Prof. Phalguni Gupta for their valuable hints and comments on the development of this research.